\documentclass[preprint]{vgtc}               % preprint
\ifpdf%                                % if we use pdflatex
  \pdfoutput=1\relax                   % create PDFs from pdfLaTeX
  \usepackage{graphicx}                % allow us to embed graphics files
  \DeclareGraphicsExtensions{.pdf,.png,.jpg,.jpeg} % for pdflatex we expect .pdf, .png, or .jpg files
\else%                                 % else we use pure latex
  \usepackage{graphicx}                % allow us to embed graphics files
  \DeclareGraphicsExtensions{.eps}     % for pure latex we expect eps files
\fi%

\graphicspath{{figures/}{pictures/}{images/}{./}} % where to search for the images

\usepackage{microtype}                 % use micro-typography (slightly more compact, better to read)
\PassOptionsToPackage{warn}{textcomp}  % to address font issues with \textrightarrow
\usepackage{textcomp}                  % use better special symbols
\usepackage{mathptmx}                  % use matching math font
\usepackage{times}                     % we use Times as the main font
\usepackage{cite}                      % needed to automatically sort the references
\usepackage{tabu}                      % only used for the table example
\usepackage{booktabs}                  % only used for the table example
\usepackage{amssymb}
\usepackage{amsmath} 
\usepackage{csquotes}
\usepackage[hyphens]{url}
\usepackage{enumitem}
\usepackage{tikz}
\usetikzlibrary{trees}
\usetikzlibrary{arrows}
\tikzset{
    >=stealth,
    auto,
    node distance=3.5cm,
    font=\scriptsize,
    cell/.style={rectangle,draw,thick,align=center},
    io/.style={double,circle,draw,thick,align=center},
    minimum size=40pt
}

\usepackage[normalem]{ulem}
\definecolor{mred}{rgb}{.80,.12,.30}
\definecolor{grey}{rgb}{0.5,0.5,0.5}
\definecolor{Purple}{rgb}{.75,0,.85}

\newif\ifnotes
\notestrue

\onlineid{0}

\vgtccategory{Research}

\vgtcinsertpkg

\preprinttext{To appear in IEEE Computer Graphics and Applications.}

\title{RNNbow: Visualizing Learning via Backpropagation Gradients in Recurrent Neural Networks}
\author{Dylan Cashman\thanks{e-mail: dylan.cashman@tufts.edu}\\ %
        \scriptsize Tufts University %
\and Genevieve Patterson\thanks{e-mail:gen@microsoft.com}\\ %
        \scriptsize Microsoft Research %
\and Abigail Mosca\thanks{e-mail: abigail.mosca@tufts.edu}\\ %
        \scriptsize Tufts University %
\and Nathan Watts\thanks{e-mail: nathan.watts@tufts.edu}\\ %
        \scriptsize Tufts University %
\and Shannon Robinson\thanks{e-mail: shannon.robinson@tufts.edu}\\ %
        \scriptsize Tufts University %        
\and Remco Chang\thanks{e-mail:remco@cs.tufts.edu}\\ %
     \scriptsize Tufts University}

\abstract{We present RNNbow, an interactive tool for visualizing the gradient
flow during backpropagation training in recurrent neural networks.  RNNbow is a web application
that displays the relative gradient contributions from Recurrent Neural Network (RNN) cells in 
a neighborhood of an element of a sequence.  We describe the calculation of backpropagation through time (BPTT) that keeps track of \textit{itemized gradients}, or gradient contributions from one element of a sequence to previous elements of a sequence.  By visualizing the gradient, as opposed to activations, RNNbow offers insight into \textit{how} the network is learning.  We use it to
explore the learning of an RNN that is trained to generate code in the 
$C$ programming language.  We show how it uncovers insights into the
\textit{vanishing gradient} as well as the evolution of training as the RNN works its way through a corpus.%
} % end of abstract

\CCScatlist{ 
  \CCScat{I.2.6}{Computing Methodologies}%
{Artificial Intelligence}{Learning};
  \CCScat{G.1.6}{Mathematics of Computing}{Numerical Analysis}{Optimization};
  \CCScat{H.1.2}{Information Systems}{Models and Principles}{User/Machine Systems}
}

\begin{document}

%% The ``\maketitle'' command must be the first command after the
%% ``\begin{document}'' command. It prepares and prints the title block.

%% the only exception to this rule is the \firstsection command
\firstsection{Introduction}
% \makeatletter
% \let\@oldmaketitle\@maketitle% Store \@maketitle
% \renewcommand{\@maketitle}{\@oldmaketitle% Update \@maketitle to insert...
%   \includegraphics[width=\linewidth,height=4\baselineskip]
%     {title_image}\bigskip}% ... an image
% \makeatother
\maketitle

\copyrighttext{
  \copyright 2018 IEEE. Personal use of this material is permitted.  Permission from IEEE must be obtained for all other uses, in any current or future media, including reprinting/republishing this material for advertising or promotional purposes, creating new collective works, for resale or redistribution to servers or lists, or reuse of any copyrighted component of this work in other works.} 
  \vspace{10mm}

Artifical Neural Networks (ANNs) have made revolutionary improvements in
classification in many domains, with particular attention given to their
ability to classify images using convolutional filters\cite{NIPS2012_4824}.  A commonly-cited 
issue with all ANNs is that they act as a black box, with large numbers of
hidden layers each individually learning their own weights resulting in a massive parameter space.  Problems of interpretability are compounded by non-linear
transformations which obfuscate interactions between each layer.  Visualizations of 
activations within convolutional neural networks have seen some success in 
illuminating the inner workings of networks to both help understanding and to assist
in hyperparameter settings\cite{DBLP:journals/corr/YosinskiCNFL15}.  However, such activation visualizations are specific to the domain of image processing, and primarily offer insight into how a network is functioning after training.  In this work, we present RNNbow, a tool for providing insight into the training of Recurrent Neural Networks (RNNs).  RNNbow visualizes values calculated during training in order to show the user if their network is learning long-term time dependencies over sequences, a necessary feature in most sequential models.  It can be used to uncover problems with poorly parameterized networks early in training.

A key insight that differentiates this work from other visualizations for deep learning is that it visualizes the \textit{gradients}, not the \textit{activations}.  Activations are the responses of the network during inference - when fed an input, what neurons are firing?  While this is instructive in comprehending how the network \textit{makes decisions}, it offers little insight into how the network \textit{learns}.  Learning in ANNs is typically accomplished via \textit{gradient descent}, a method which minimizes loss over a training set by iteratively updating parameters in the direction dictated by the gradient of that loss.  Thus, to analyze how the network is learning (or if it is learning at all), we must inspect the gradients.

\begin{figure}[ht!]
 \includegraphics[width=0.5\textwidth]{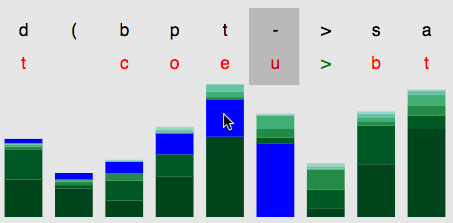}
 \caption{RNNbow helps the user see the flow of gradients due to an individual cell's loss during training of a Recurrent Neural Network.  Here, we see highlighted in blue the gradient resulting from loss due to predicting the character ``u'' when the true character was ``-''.}
 \label{ref:title_image}
\end{figure}

RNNs are a particular class of ANNs that map input sequences to output sequences.  As with all ANNs, their function depends on what they are fed in as inputs and what they are fed as desired outputs.  They can learn to label each item in a sequence if their training data includes labels; a good example of this is training an RNN to do part-of-speech tagging.  Alternatively, RNNs can be used to generate sequences that look like the training data.  This is accomplished in a technique first proposed by Elman\cite{elman1990finding} in which, for a given training set $s_1\ldots s_n$ the input sequence is set to $s_1\ldots s_{n-1}$ and the output sequence is set to $s_2 \ldots s_n$.  In this way, the RNN learns to predict the succeeding element of a sequence.
RNNs are behind some of Deep Learning's most astonishing results,
including language translation, generating image captions, and predicting medical
outcomes.  

RNNs have been called both ``unreasonably effective''\cite{unreasonable_effectiveness_rnn} and ``difficult to train''\cite{Pascanu:2013:DTR:3042817.3043083}.  One of the major issues with training RNNs is ensuring that the gradient descent updates propagate far enough back that long-term dependencies can be learned.  Consider an RNN that tried to produce the following sentence.

% \begin{displayquote}
%  \textit{The }\textbf{man} \textit{receieved a phone call on the way to} \textbf{his} \textit{car.}
% \end{displayquote}

\begin{figure}[ht!]
\includegraphics[width=0.43\textwidth]{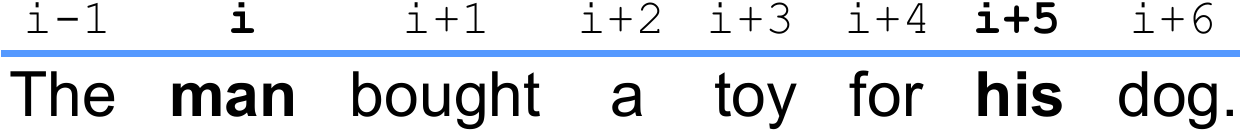}
 \label{ref:mandog}
\end{figure}

In order for the RNN to be able to know the gender of the pronoun \textbf{his}, it must remember the gendered noun \textbf{man} 5 time steps earlier.  Since RNNs learn via gradient descent, the only way to learn time dependencies of that distance is to have the gradient due to the loss incurred by an errant prediction propagate back to update the parameters that controlled how much the model learned at an earlier step.  In other words, if the word \textbf{his} at $t=i+5$ is a function of the word \textbf{man} at $t=i$, then the gradient at $t=i$ with respect to the loss at $t=i+5$ must be greater than 0.

If an RNN is parameterized poorly, it may fall victim to the well-studied \textit{vanishing gradient} problem\cite{Pascanu:2013:DTR:3042817.3043083}, in which gradient only flows a few cells back, at which point the network may be no more capable than using frequency counts over the training set.  To try to address this problem, the user must not only select numerical parameters like the size of the hidden layer or the number of layers, but also must choose between different architectures (stacks or grids) and different RNN cell types such as Long Short-Term Memory cells (LSTMs) or Gated Recurrent Units (GRUs).  The theoretical distinctions involved in making these choices can be mystifying to many users of RNNs, and it can be confusing to try to diagnose learning issues resulting from poor RNN design.  Tools are needed to reveal endemic issues in gradient flow in RNNs so that the user has early evidence of whether their network architecture is able to learn or not.

RNNbow is a tool to visualize the gradient flow during training of an RNN.  It provides an overview of the magnitude of the gradient updates thorughout training, showing the user how quickly a model is learning, and how the regime of parameter updates changes over the course of training.  It also allows users to drill down into a particular batch and view the individual influence of each of the RNN's predictions on on parameter updates in the hidden layer.  By breaking down the gradient update at each cell by each component's origin, it makes the vanishing gradient apparent.  It helps users assess their parameterization of their network during training.  It also provides an illustration of the change in gradient behavior as a network trains.  In the case of the earlier example, RNNbow can help the user detect if the loss incurred during the update of the word \textbf{him} is successfully propagated 5 steps back to the word \textbf{man}.  At the time of writing, it is the only neural network visualization that visualizes gradient flow in RNNs that the authors are aware of. 
% It may also be useful as an educational tool, as it makes mathematical phenomena more apparent to the casual observer.  

% \strike{In contrast to many of the prevalent ANN visualizations that focus on convolutional neural networks that operate on images, it is agnostic to the domain - while it is demonstrated in the paper as working on a character-level RNN, it could be modified to show video frames or words instead of characters.}

Further, because RNNbow visualizes the gradient and not the input space, the use of the tool is agnostic to the domain of the problem. In contrast to many of the prevalent ANN visualizations that focus on convolutional neural networks that operate on images\cite{DBLP:journals/corr/LiuSLLZL16,DBLP:journals/corr/YosinskiCNFL15}, RNNbow could be used to visualize the gradient of any RNN. In this paper we use a character-level RNN as a demonstration, but RNNbow could be applied to show learning of other sequential data, including video frames and words.

For a use case, we repeat a well-known RNN experiment\cite{DBLP:journals/corr/KarpathyJL15} to learn and generate statements in the C programming language via a character-level RNN.  We present some insights that can be gleaned via RNNbow.  We explain how traditional implementations of backpropagation can be modified to collect the itemized gradients visualized by RNNbow, and discuss complexity implications.  We discuss the advantages of visualizing gradient over activation, discuss the role of visual analytics in deep learning, and conclude by considering future work in using RNNbow to compare different architectures.

%% \section{Introduction} %for journal use above \firstsection{..} instead
% Artificial Neural Networks have become the flagship statistical tool for 
% classification, making drastic improvements in competitions for image
% classification\cite{NIPS2012_4824}, video tagging\cite{DBLP:journals/corr/NgHVVMT15}, 
% and many other domains, generally making use of convolutions to exploit 
% locality in the data.  More recently, Recurrent Neural Networks have made 
% public breakthroughs in sentence translation\cite{DBLP:journals/corr/WuSCLNMKCGMKSJL16}, 
% image captioning\cite{Karpathy:2017:DVA:3069214.3069250} Recurrent Neural 
% Networks are a particular class of ANNs that represent sequences, such as 
% frames in a movie

\section{Related Work}

Prior to the recent explosion in big-data neural networks, artificial neural networks were generally small enough to allow for overview visualizations of all nodes and edges in their computation graph.  Early work in visualizing neural network activity focused on ``opening the black box'' in this complicated computation graph.  A good example can be found in Tzeng and Ma's work to display three-layer networks as node-link diagrams, using the size and color of the nodes and edges to encode activation magnitude and uncertainty\cite{tzeng2005opening}.  As the number of layers increases, such visualizations did not scale, and visualizations began to focus on either aggregate views of activations on particular inputs, or by viewing inputs that maximize activation of particular nodes\cite{DBLP:journals/corr/YosinskiCNFL15}.  Some visualizations of the popular Convolutional Neural Network take advantage of the visual form of the input space, integrating images into overviews of the node activations\cite{DBLP:journals/corr/LiuSLLZL16}.  Some visualizations treat neural networks like other similar high-dimensional classifiers, visualizing 2-D projections of their classifications to provide insight into their decision boundaries\cite{kahng2018cti}.

Because of their sequential nature, RNNs proffer an opportunity for more concrete temporal visualizations.  In an influential blog post and accompanying publication\cite{DBLP:journals/corr/KarpathyJL15}, Karpathy et al. used a variety of visualizations that overlayed some representation of node activation over subsets of the input space to show how different hidden nodes are responsible for different decision logic.  There has also been work in interpreting the hidden state dynamics of a trained RNN \cite{8017583,DBLP:journals/corr/abs-1710-10777}. Their visualizations suggested that activations in the hidden layer contain information about the length of memory in a model.

Most of the tools listed are used after training to attempt to render interpretable the state of a trained network at test time.  In contrast, RNNbow is used to visualize how the network has learned.  Thus, it could be useful to view gradients during training, to know whether hyperparameters need to evolve or if the experiment needs to be rerun with a different set of hyperparameters.  Mid-training visualization is one of the features of \textit{TensorBoard}, a visualization tool built on top of Google's \textit{TensorFlow}\cite{tensorflow2015-whitepaper}.  \textit{TensorBoard} allows users to write out values calculated during training to a log, and then generates basic visualizations, such as line graphs and bar charts, of those values throughout training.  A typical use case is to plot the loss of a network as a function of the number of batches trained to confirm that the model is improving performance.  While it would be possible to log gradients by patching the backpropagation calculation in a \textit{TensorFlow} project, there is minimal support for visualizing those gradients beyond line graphs and bar charts at the time of writing.  A recent work by Liu et. al. does visualize the training process for deep generative models \cite{8019879} by plotting the data flow of activations through layers accompanied by basic measures of performance such as accuracy over time.

Our tool is unique in several ways.  First, it visualizes gradient flow as opposed to activations.  RNNbow is also used to assess the learning of a network during training, to determine if a change in hyperparameters is needed, as opposed to analyzing an ANN's response during test time.  Lastly, in contrast to the cited RNN visualizations, RNNbow is agnostic to type of input sequence (text characters, movie frames, medical records) since it does not use the input domain as a fundamental part of the visualization.

% I like using the below as a way of classifying previous visualizations and showing where this one sits.
% There are several different types of visualizations that are useful for complicated machine learning methods like neural networks.

% \begin{itemize}
% \item \textbf{Aiding model customization. } Visualizations can be used to help the user choose hyperparameters for the model.  This type of visualization is more useful for an expert who understands the significance of all chosen paremeters.  This is generally done prior to or during training.
% \item \textbf{Educational aid. } Visualizations of machine learning methods can be helpful in educating users of the generic use of those methods.
% \item \textbf{Post-hoc output analysis. } Many neural network visualizations are used to try to grasp just what an already-trained network is doing.  
% \end{itemize}

\begin{figure}[ht!]
  \centering
  \includegraphics[scale=0.36]{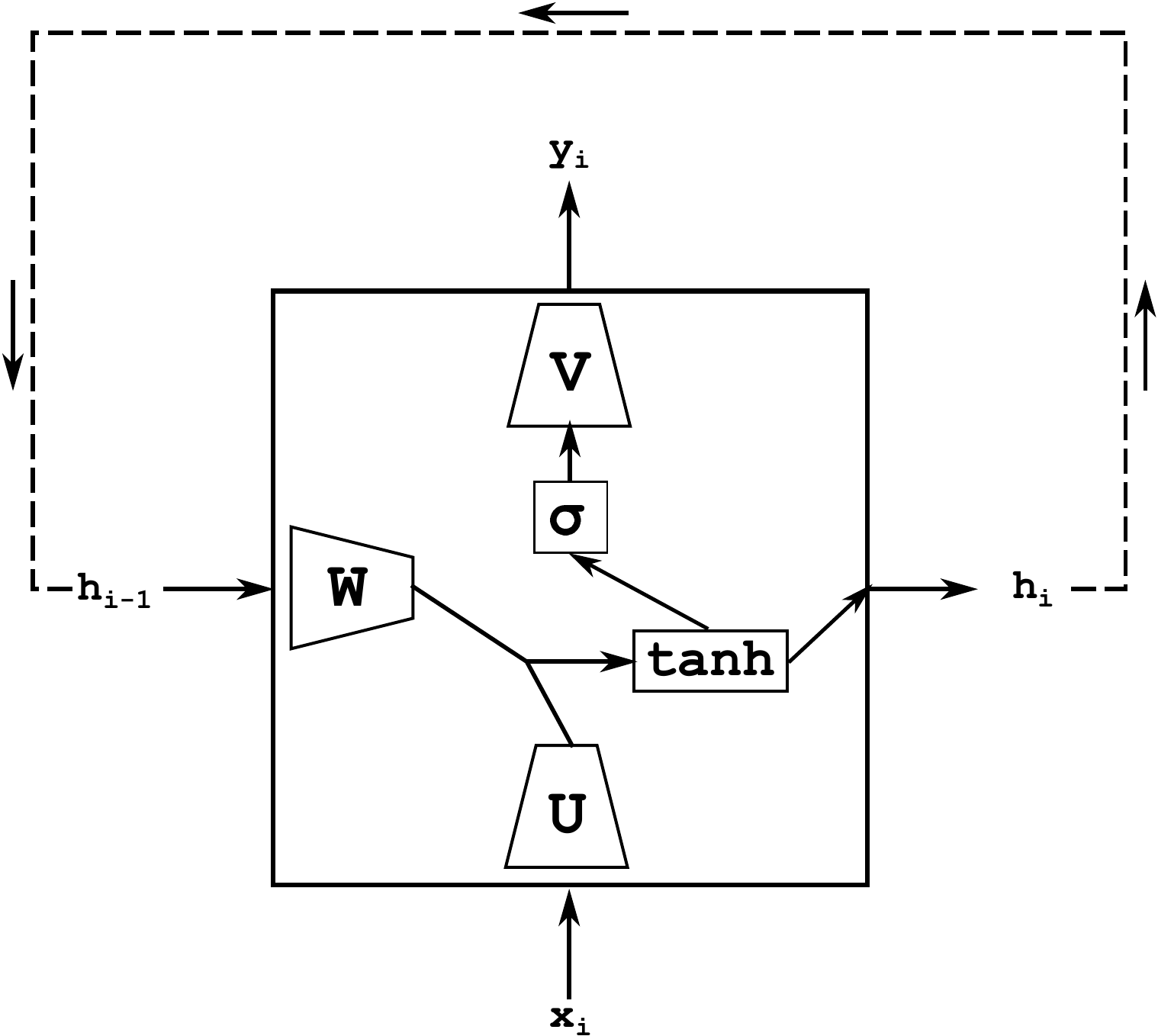}

  \caption{A simple one-cell recurrent neural network, seen as a cyclic computation graph.  Trapezoids are linear transformations by a weight matrix.  Rectangles are element-wise scalar functions.  The RNN produces an output $y_i$ for every input $x_i$, passing on the calculated hidden state $h_i$ back to itself to use for the next input.}
  \label{ref:cyclic}
\end{figure}

\begin{figure}[ht!]
  \centering
  \includegraphics[scale=0.29]{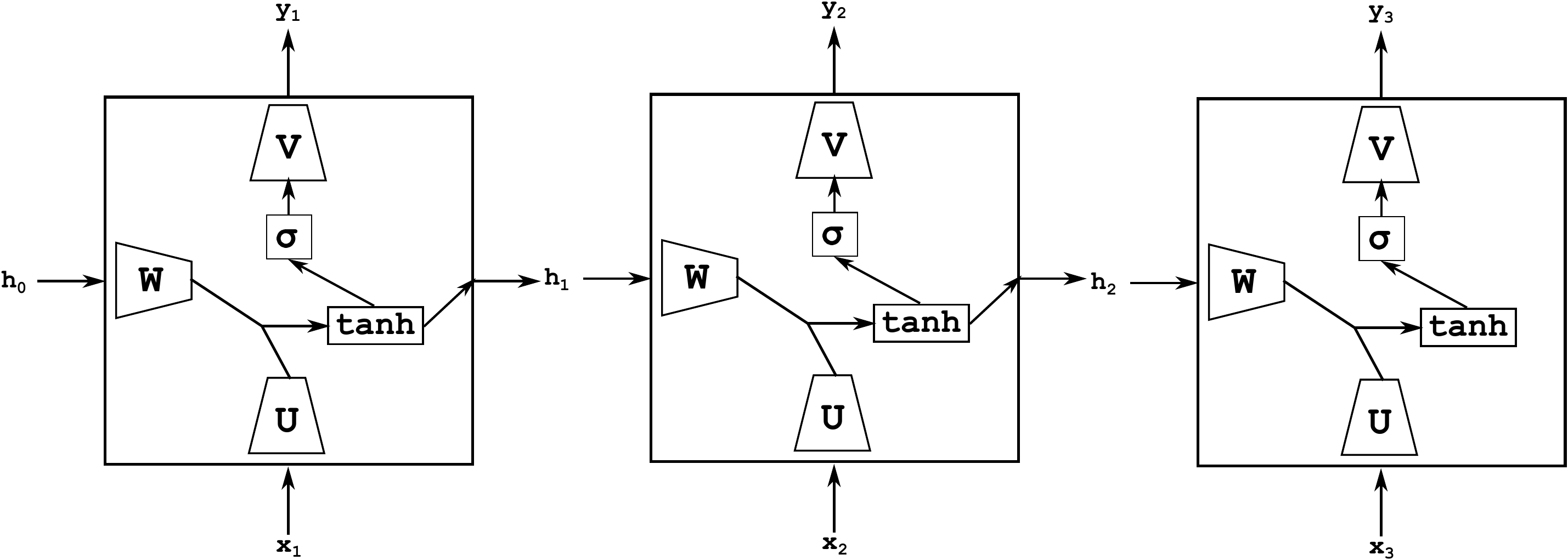}
  \caption{To make calculation over an input sequence well-defined, the RNN from Fig~\ref{ref:cyclic} is unrolled once for each element in the input sequence.  This RNN is unrolled three times to make three cells.  It takes three inputs, and produces three outputs.  The weight matrices $U$, $W$, and $V$ are shared in each cell. }
  \label{ref:unrolled}
\end{figure}

\section{Recurrent Neural Networks}

The goal of an RNN is to produce output given sequence input.  Their advantage over other sequential learners such as Markov chains or Maximum Entropy Classifiers is that they are able to learn long-term dependencies via nonlinear dynamics in their hidden layer.  The basic RNN architecture can be viewed as a graph with cycles, as seen in Figure \ref{ref:cyclic}.  At any given point in inference, the input $x_i$ and the previous hidden state $h_{i-1}$ are used to calculate the new hidden state $h_i$, which is then used to calculate the emission $y_i$. Mathematically, this can be described as:

\begin{align}
h_i &= \tanh{(W h_{i-1} + U x_i)} \label{eqn:hidden_state}\\
y_i &= \sigma (V h_i)\label{eqn:output_calc}
\end{align}

Here, $W$, $U$, and $V$ are weight matrices, and $\sigma$ is the sigmoid function $\sigma (x) =  \frac{1}{1 - e^{-x}}$.  Both $\tanh$ and $\sigma$ are common \textit{activation functions} in the deep learning literature.  Intuitively, the weight matrices perform a linear transformation on the data, and then the activation functions squash the values back to an interpretable, normalized range, with $\tanh$ bounded by $(-1,1)$, and $\sigma$ bounded by $(0,1)$.  In addition, these activation functions add a nonlinearity into computation so that the RNN can fit more than polynomial functions.

During training, the training data set is partitioned into regular \textit{batches}.  An RNN trains on one batch at a time, in sequential order, by unrolling for a number of steps equal to the size of the batch.  A batch size of 3 is seen in Figure~\ref{ref:unrolled}; however, a typical batch size might range from a dozen elements to around a hundred.  Within a batch, the RNN steps through input in order, taking in an input, calculating a hidden state, emitting an output, and then passing on the hidden state to be used for the next item in the sequence.  The inputs are any data that can be sequenced (characters, words, frames, etc.), and the outputs can be classifications of or regression on those inputs, or distributions of potential classifications over the output range.  The outputs are compared to the true labels, and a loss is calculated.  The objective of training is to minimize the loss by choosing the optimal weight matrices $W$, $U$, and $V$.  After total loss has been calculated for an entire batch, the gradient of the loss with respect to these weight matrices is calculated and they are then updated via gradient descent.  These gradients are typically calculated using \textit{backpropagation}, an efficient algorithm for calculating gradients in computational graphs.  The newly updated weight matrices are used for the next batch.  The batch size, the size of the hidden layer, and certain constants used in the updating of the weight matrices are all hyperparameters set by the user.

For example, in our use case described in section \ref{sec:use_case}, we build an RNN to generate code for the \textit{C} programming language.  Before training, $W$, $U$, and $V$ are initialized randomly.  If we used our RNN with these randomly-initialized weight matrices to generate text, it would be the same as sampling from a uniform distribution over all characters, and thus it would not look like code.  By feeding our RNN input data that looks like valid code, we gradually update our weight matrices so that our RNN generates sequences that better match not only the distribution of characters in the training set but the transitions between characters as well.  Examples of code generated by this RNN before and after training can be seen in Figure~\ref{fig:generated_text}.  For more examples of character level RNNs, including much more in-depth analysis of the generation of $C$ code, see \cite{DBLP:journals/corr/KarpathyJL15}.

\begin{figure}[ht!]
\hspace{6mm}Batch 0: \hspace{6mm}\verb#0cu    |nv"M$R/m^u tt+^CeU@x>Uh#

\hspace{6mm}Batch 10000: \hspace{6mm}\verb|s ged->bat ag_Me_sertaket())|

\caption{Text generated by a character level RNN as used in our use case.  The first line was generated before any training, and seems like a random sampling of characters.  The second line was generated after training on 250000 characters of the Linux Kernel, and seems to have captured some understanding of the syntactic rules of the \textit{C} programming language, such as function calls, underscore separators in function names, and pointer accessing. }

\label{fig:generated_text}
\end{figure}

Our full training dataset is the Linux kernel, which we split into batches of 25 characters.  This also corresponds to unrolling our RNN 25 steps.  Referring to the equations defined in (\ref{eqn:hidden_state}) and (\ref{eqn:output_calc}), the character input $x_i$ could be encoded densely using a mapping such as ASCII, or it could use one-hot encoding.  We use a hidden layer of 100 nodes, meaning that the weight matrix $U$ transforms our input $x_i$ into a 100-dimensional vector representation of the original input character.  The hidden state vector is also 100 dimensions.  $Wh_{i-1}$ and $Ux_i$ are added together and then squashed through the $\tanh$ activation function.  That result is then multiplied by the weight matrix $V$, which projects back into the character-space encoding, providing a multinomial distribution over all possible characters.

\begin{figure*}[ht!]
  \includegraphics[width=\textwidth]{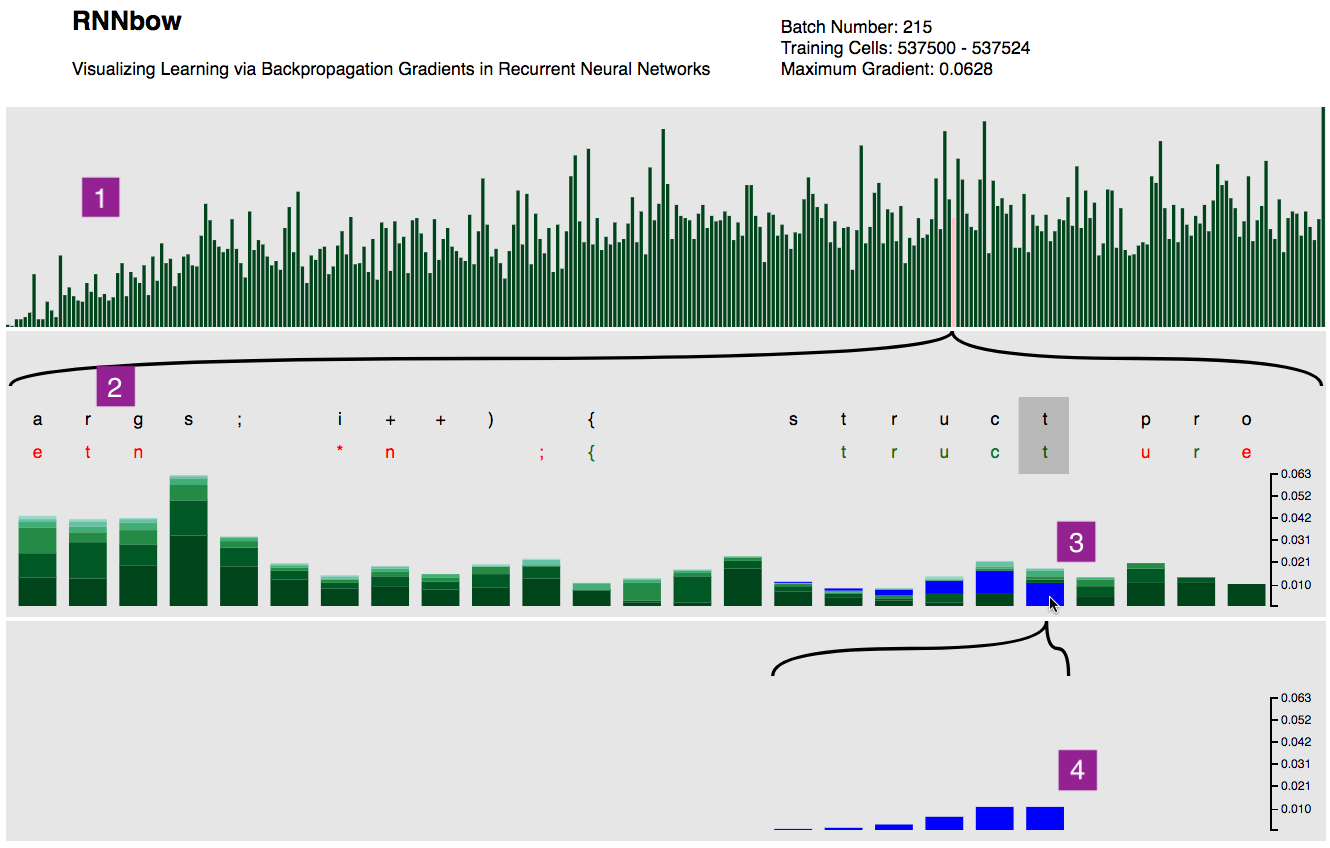}
  \caption{The user interface to RNNbow.  In (1), the user is shown a bar chart where each bar represents the maximum gradient per batch.  By mousing over different batches, the user can drill down to view gradient data from each batch in the training set.  The pink bar seen $3/4$ of the way through (1) indicates the currently selected batch, and some information for that batch is seen printed above (1).  In (2), the top row of characters holds the true labels from the training set, and the bottom row holds the prediction from the RNN at training time.  The prediction is colored green if it is correct, and red if it is incorrect.  (3) shows the magnitude of the gradients being used to update the weights at each point in time.  Each bar is decomposed into the different sources of the loss that created that gradient.  On mousing over a particular gradient, we see the gradient due to a single loss highlighted in blue, and projected down to (4) for easy inspection.    Different batches can be previewed in (2), (3), and (4) by hovering over their respective bars in (1), and selected by clicking on those bars.}
  \label{ref:interface}
\end{figure*}

To train on a given batch, the RNN starts with the first character as input, calculates a hidden state, and outputs a multinomial distribution for what it thinks the next character could be.  In RNNbow, we show the most-likely character from that multinomial distribution, $\max(y_i)$, as shown in area 2 of Figure~\ref{ref:interface}.  The true label for the first character in the batch is the second character in the batch, since in this experiment, we want the RNN to predict succeeding characters based on the current character and its context.  We use a softmax loss, which generates high loss if our RNN suggests there is a low probability of the true label and a high probability of other labels.  That loss is then used to calculate the gradient update for the weight matrices.  
% I don't think this sentence should go here, but it has to go somewhere.
In RNNbow, we only visualize the gradients of $W$, since $W$ is what controls the memory of the RNN.

\section{RNNbow}

RNNbow is a web application that visualizes the gradients used to update parameters during training of a recurrent neural network.  In this section, we describe the interface, then we review how the gradient data is harvested during training via backpropagation through time (BPTT)\cite{Hochreiter:1998:VGP:353515.355233}.  

\subsection{Interface}
% \remco{this section can be organized further for ease of reading. Do either subsubsections or use begin{description} with the first word bold-faced to save space. For example:}

% \begin{description}[topsep=1ex,itemsep=0ex,partopsep=1ex,parsep=1ex,leftmargin=2ex]
% \item [User Interactions] blah blah blah

% \item [Area 1: Prediction and True Targets] blah blah blah

% \item [Area 2: Per Batch Gradients] blah blah blah

% \item [Area 3: Per Prediction Gradients] blah blah blah

% \item [Area 4: Gradients of All Batches] blah blah blah

% \end{description}

The user interface of RNNbow is a coordinated multiple view implemented in \textit{d3.js} that provides both an overview of the data as well as details of particular elements of the training set.  The interface can be seen in Figure~\ref{ref:interface}.  It takes in data on gradients recorded during a pass over a training set.  The specific data-visual mappings and how such data is generated during training of an RNN model are both described in section~\ref{sec:bptt}.  RNNbow provides both an overview of gradients across all batches and the ability to drill down and visualize the gradients from a single batch at a time, so that the gradients at individual locations in the training set can be seen clearly.  Descriptions of the interface below will make repeated use of the numeric labels from Figure~\ref{ref:interface}.  

\subsubsection{Area 1: Overview of Max Gradient of All Batches }
Area 1 of the interface is a bar chart overview of the maximum gradient within each batch.  Our data comprised of 300 batches of gradient data during training.  Each bar in area 1 represents the maximum gradient across all time steps within that batch.  Because RNNbow visualizes the maximum gradient in each batch as the height of a bar in a bar chart, the user is able to easily to navigate to parts of their training set where the most learning is happening.  While visualizing the mean would also be valid in that it would show which batches were the most informative, visualizing the max instead shows which individual elements of the training set are the most informative.

The user is able to drill down into a particular batch by hovering over a particular bar, resulting in information about that batch being visualized in areas 2, 3, and 4.  To fix that batch as the selected batch, the user may click on a bar.  The currently selected batch is signified as a pink bar in area 4.  Some basic information about the batch being visualized is displayed in text above area 1, including the batch number, which training cells it corresponded to, and the maximum gradient in that batch.
% \textbf{Area 1: Prediction and True Targets.} 
\subsubsection{Area 2: Prediction and True Labels of a Single Batch}
Area 2 provides details on demand for the batch selected in are 1.  In the top row of characters in area 2, we can see the ground truth labels from the training set per element in the batch.  Immediately underneath those labels, we see our RNN's prediction for the label of that element.  These predictions are colored according to whether they are correct (green) or incorrect (red).  In this figure, the training set batch begins with the five characters ``\texttt{args;~}'', and our RNN predicts the five characters ``\texttt{etn~~}''.  Showing the true and predicted labels helps ground the user in their data.  

% \textbf{Area 2: Per Batch Gradients. } 
\subsubsection{Area 3: Per Batch Gradients }
Underneath each label, in area 3, gradients at each time step of the selected batch of training data are visualized as a stacked bar chart.  The height of each bar represents the magnitude of the gradient used to update the weights of the RNN \textit{at that step in time}, relative to the gradients within that specific batch.  The bars are partitioned according to how far in the future that portion of the gradient resulted from.  The lowest, darkest portion of the bar is the gradient due to the loss of the current point in time (due to the loss of the label immediately above the bar).  Stepping up in the stacked bar, each new partition is gradient due to the loss accrued due to the succeeding label.  In RNNbow, each vertical bar can show the gradient contribution from up to 5 time steps in the future.  The number of steps was chosen empirically based on this use case; for other datasets, a larger horizon may be necessary.  For more discussion, see section~\ref{sec:computational_concerns}.

% For example, the total gradient is large for the fourth character in the batch shown in area 3 of Figure \ref{ref:interface}, which we'll call $t=4$.  The RNN predicted a white space character, but the true label was the character `\texttt{s}`. The darkest green portion of the bar below the character \texttt{s} represents the gradient due to that particular prediction.  However, predictions made in succeeding steps $t=5,6,\ldots$ are a function of the hidden state calculated at $t=4$, so there is also gradient in this node due to poor predictions made in the succeeding time steps.  Thus, the slightly lighter green portion of the bar beneath the \texttt{s} is the amount of gradient that is the result of the prediction made at $t=5$, where the RNN predicted a white space character but the true label was the character `\texttt{;}`.

% \textbf{Area 3: Gradients Due to Individual Prediction. } 
\subsubsection{Area 4: Gradients Due to Individual Prediction }
While seeing the breakdown of the sum of the gradient at each step may be informative, it is also interesting to see how gradient flows backwards from a particular time step - this would show how the network was learning long-term dependencies.  By hovering over any portion of a bar in area 3, we highlight all portions in neighboring bars due to the same loss.  In addition, the particular prediction and label that are responsible for that loss are highlighted with a darker gray background, as seen in the gray box behind the two ``t'' characters in Figure~\ref{ref:interface}.  For the sake of analysis, these portions are projected down into area 4 to highlight the rate at which the gradient decays.  

As an example, in Figure~\ref{ref:interface}, the cursor is hovering over the bottom component of the gradient bar below the true label `\texttt{t}`.  In area 4, we can see that the gradient due to this decision propagated back 5 time steps, albeit diminishing in magnitude.  Each blue bar represents the amount the parameters are being updated at that point in time due to the prediction made at `t`.  The magnitude of these bars is a proxy for how large the influence is of a previously predicted character, such as the white space character instead of an `s` at the beginning of `struct`, had on the prediction of the character `t` at the end of `struct`.  The faster this gradient decays, the shorter the time dependency is that the model is learning.  We call this projected bar chart the \textit{gradient horizon}, as it aims to show when the gradient contribution vanishes as it passes back.  As the user sweeps the mouse up and down and across bars, area 4 changes which gradient horizon it displays, allowing the user to quickly navigate the decomposition of gradients.

% \textbf{Area 4: Overview of Max Gradient of All Batches. } 

\subsection{Generating Gradient Data}
\label{sec:bptt}

% \strike{In this section, we explain modifications to a basic RNN implementation
% to record the gradients needed by RNNbow.  We derive the equation for 
% $\frac{\partial L}{\partial W}$ via the chain and product rules, explaining
% how to collect the visualized gradients along the way.  We show that this
% derivation is equivalent to backpropagation, and explain how we are trading
% off computational efficiency for the ability to record individual gradients.}

Given a loss function, backpropagation passes that loss back to any parameters involved in the loss's calculation, via the chain rule.  In a Convolutional Neural Network, where one prediction is made, there is generally a single loss calculation, which is then passed along via gradients.  In an RNN, there are multiple losses; loss is calculated at each output $y_i$.  Calculating the gradient of $W$ is a difficult task since each hidden state and each output are compounded functions of $W$.  

To account for the multiple losses, RNNs use a special form of an algorithm
called backpropagation through time (BPTT)\cite{Hochreiter:1998:VGP:353515.355233}.  To use BPTT, RNNs are \textit{unrolled} - that is, each 
cycle in the computation graph is represented as an additional copy of the RNN, to create
a directed acyclic computation graph that backpropagation can then be used on, as seen in Figure~\ref{ref:unrolled}.

Backpropagation is designed to be as computationally fast as possible, making extensive use of dynamic programming to memoize intermediate calculations so that the gradient can be calculated in a single pass backwards through time.  However, fully utilizing dynamic programming will cause us to lose track of some of the intra-sequence effects that RNNbow aims to illuminate.  Thus, we remove one level of dynamic programming, trading off increased computational complexity for the ability to record itemized gradients.  To motivate this, in the following section we fully derive an expression for the gradient, pointing out what the mapping is between RNNbow and the terms in that expression.  We also show how our implementation is equivalent to backpropagation.

\subsubsection{Derivation of Itemized Gradients}

We are concerned with $\frac{\partial L}{\partial W}$, the rate at which the loss ($L$) changes with respect to the weights of the hidden layer ($W$).  In training, we use that quantity to update $W$ via gradient descent.  Because we are interested in the gradient contributions from each time step, we decompose the loss into loss contributed from the prediction made at each time step.  Here, $L_t$ is defined as the loss due to predicting $y_t$, and $n$ is the size of the batch.

\begin{align}
L &= \sum_{t=1}^n L_t\\
\frac{\partial L}{\partial W} &= \sum_{t=1}^n \frac{\partial L_t}{\partial W}\label{eqn:sum_loss_grad}
\end{align}

For a given time step $t=i$, we have the following decomposition, via the chain rule.

% \begin{equation}
% \frac{\partial L_i}{\partial W} = 
% \frac{\partial L_i}{\partial y_i} \cdot 
% \frac{\partial y_i}{\partial W} + 
% \frac{\partial L_i}{\partial h_i} \cdot 
% \frac{\partial h_i}{\partial W}
% \end{equation}

% The gradient descent update at time step $t=i$ can be decomposed into gradient $\frac{\partial L}{\partial y_i} \cdot \frac{\partial y_i}{\partial W}$ due to the loss at $t=i$, and gradient that flows back along the hidden layer from later points in time, $\frac{\partial L}{\partial h_i} \cdot \frac{\partial h_i}{\partial W}$.

\begin{align}
\frac{\partial L_i}{\partial W} &= \frac{\partial L_i}{\partial y_i} \cdot \frac{\partial y_i}{\partial W} \nonumber \\
&= \frac{\partial L_i}{\partial y_i} \cdot \frac{\partial y_i}{\partial h_i} \cdot \frac{\partial h_i}{\partial W} \label{eqn:naive_grad_i}
\end{align}

We can further decompose the rightmost term, $\frac{\partial h_i}{\partial W}$, but we must be careful: $h_i$ is a function of $W$, but it is also a function of $h_{i-1}$ which is in turn a function of $W$, so we must use the product rule.

\begin{align}
\frac{\partial h_i}{\partial W} &= \tanh'(Ux_i + Wh_{i-1}) \left[ h_{i-1} + W\frac{\partial h_{i-1}}{\partial W}\right]
\end{align}

The leftmost term is the derivative of $\tanh(x)$, evaluated at $Ux_i + Wh_{i-1}$.  Notice that we still must further expand the rightmost term, just like we had to with $\frac{\partial h_i}{\partial W}$.  In the following derivations, the term $\tanh'(Ux_i + Wh_{i-1})$ is truncated to $\tanh'_i$ for the sake of readability.

\begin{align}
\frac{\partial h_i}{\partial W} &= \tanh'_i \left[ h_{i-1} + W\tanh'_{i-1}\left[ h_{i-2}  + W \frac{\partial h_{i-2}}{\partial W}\right]\right]
\end{align}

We would then have to expand $\frac{\partial h_{i-2}}{\partial W}$, and $\frac{\partial h_{i-3}}{\partial W}$, and on until we end up with the term $\frac{\partial h_{1}}{\partial W}$ which does not expand since $h_0$, our initialized hidden state, does not depend on $W$ - it is a hyperparameter set by the user.  In this way, the loss due to our prediction at $t=i$ ends up propagating all the way back through the input sequence to $t=1$.

\begin{align}
\frac{\partial h_i}{\partial W} &= \tanh'_i \left[ h_{i-1} + W\tanh'_{i-1}\left[ h_{i-2}  + \ldots + W \frac{\partial h_{1}}{\partial W}\right]\right] \label{eqn:full_grad_sum}
\end{align}

If we were to expand out all of the products in (\ref{eqn:full_grad_sum}), we would end up with summands that were only dependent on values available in ordered subsets of the sequence.

\begin{align}
\frac{\partial h_i}{\partial W} &= \tanh'_i h_{i-1} + \tanh'_iW\tanh'_{i-1} h_{i-2}  + \ldots \label{eqn:expanded_grad_sum}
\end{align}

Note that calculating the first summand only requires knowing $h_{i-1}$ and the additional arguments to $\tanh'_i$, $U$, $W$, and $x_i$.  Then, we can memoize the value of $\tanh'_i$, and when calculating the next summand, we only need that memoized value and $U$, $W$, $h_{i-2}$, and $x_{i-1}$.  Let $M_j$ be the $i-j$th summand of (\ref{eqn:expanded_grad_sum}), so $M_i = \tanh'_i h_{i-1}$, $M_{i-1} = \tanh'_iW\tanh'_{i-1} h_{i-2}$.  Note that calculating $M_j$ depends only on the values $U$, $W$, $x_j$, $h_{j-1}$, $h_j$, and $M_{j+1}$. 

% \remco{I'm a little lost here with the $M_j$ and $M_i$... }

\begin{align}
\frac{\partial h_i}{\partial W} &= M_i + M_{i-1} + \ldots + M_1 \label{eqn:m_decomp}\\
M_j &= \frac{M_{j+1}}{h_j} W \tanh'_{j}h_{j-1} \hspace{5mm} ; \hspace{5mm} 0 < j < i \label{eqn:m_ratio}\\
M_i &= \tanh'_i h_{i-1}
\end{align}

Then we can rewrite (\ref{eqn:naive_grad_i}) in a way that clarifies our implementation of its calculation.

\begin{align}
\frac{\partial L_i}{\partial W} &= \sum_{j=1}^i \frac{\partial L_i}{\partial y_i} \cdot \frac{\partial y_i}{\partial h_i} \cdot M_j \label{eqn:naive_expanded_grad_i}
\end{align}

In order to calculate the gradient for the entire batch, we would need to sum this over each time step, so we substitute (\ref{eqn:naive_expanded_grad_i}) into (\ref{eqn:sum_loss_grad}).

\begin{align}
\frac{\partial L}{\partial W} &= \sum_{t=1}^n \sum_{j=1}^t\frac{\partial L_t}{\partial y_t} \cdot \frac{\partial y_t}{\partial h_t} \cdot M_j \label{eqn:full_grad_double_sum}
\end{align}

In order to use RNNbow, we record the summand of (\ref{eqn:full_grad_double_sum}) for each value of $(t,j)$.  We call these summands the \textit{itemized gradients}, as they are itemized by the time step that was the source of their loss.

To calculate this in $O(n^2), \{M_j\}$ can be implemented as a one-dimensional dynamic programming table that is filled in from right to left.  Then each  can be calculated in a single backward pass of the batch, from $j=t$ down to $j = 1$.  As an example, suppose that we were training a character-level RNN, and had a batch to train on that was the six characters \texttt{g u i t a r}, but our RNN instead predicted the six characters \texttt{b a n a n a}.  
% This is illustrated in Figure~\ref{img:bptt_algo}.  
To calculate our gradient, we start at the last character, $t=6$, and see that we predicted \textit{a} instead of \textit{r}, and so we incorporate some loss.  We record the gradient due to that prediction at $t=6$, and then pass back that loss via $M_6$ to $t=5, \cdots, t=1$.  Once we have calculated all itemized gradients due to predicting \textit{a} instead of \textit{r} at $t=6$, we jump to $t=5$, calculate the loss due to predicting \textit{n} instead of \textit{a}, and pass a different set of $M_j$ back.  This is based on an implementation of BPTT from \cite{wild_ml_rnn3}.  

% \begin{figure}[ht!]
%  \label{img:bptt_algo}
%  \includegraphics[width=0.5\textwidth]{rnnvis_imgs/bptt_draft}
%  \caption{The modified version of BPTT used to save itemized gradients.  This illustration shows the calculation of all contributions to the gradient from the loss at time $t=6$ when trying to predict the word \textit{guitar}.}
% \end{figure}

The full area of the batch view seen in area 3 of Figure~\ref{ref:interface} represents the full value of $\frac{\partial L}{\partial W}$.  The full area of the detailed gradient horizon seen in area 4 of Figure~\ref{ref:interface} represents the full value of $\frac{\partial L_i}{\partial W}$ described in (\ref{eqn:naive_expanded_grad_i}), and each bar within area 3 corresponds to an individual summand.  A vanishing gradient would correspond to the summands decreasing as $j$ decreases, which can be seen in Figure~\ref{img:vanishing_gradient}.

Traditional backpropagation only takes $O(n)$, but it doesn't expose the itemized gradients that we need to record for RNNbow.  A further exploration of the relationship between our calculation and traditional backpropagation is given in the appendix.

% Generally, the gradient descent update is calculated in a single pass backwards through time, passing back the gradient at each step.  While this is computationally efficient, we actually lose track of the individual contributions due to each time steps loss - they all get absorbed into the gradient passed back. Thus, to gather the itemized gradients visualized by RNNbow, we have to modify this computation.  Instead of calculating both gradient due to $y_i$ \textit{and} gradient due to $h_i$, we only calculate the gradient due to $y_i$.  Then, we step back in time for $k$ steps, passing that individual loss's gradient back.  After $k$ steps, we then progress to $t=i-1$.  In our use case, we set $k=5$.

\subsubsection{Computational Concerns vs. Estimation}
\label{sec:computational_concerns}
In practice, it may be impractical and unadvised to calculate the itemized gradients throughout all of training.  To begin with, this algorithm generates an immense amount of data, storing $O(HNn)$ gradients in a single pass over the training set, where $H$ is the number of nodes in the hidden layer, $N$ is the size of the training set, and $n$ is the batch size.  In our use case, we used small batches and a small hidden layer ($n=25, H=100$) compared to many networks, and if we had used the entire training set, even using these small settings for $n$ and $H$ we would have created data that was 2500 times the size of our training data.  

The problem of data size can be ameliorated by only calculating the itemized gradients periodically - in our use case, we only store the gradients every 100 batches, reverting to the optimized version of backpropagation for the other 99\% of batches.  Lastly, gradients are averaged between all nodes in the hidden layer, as our goal is to see the general rate of training, rather than drilling down into individual hidden nodes.  Since this data is used for visualizations, visualizing data from all nodes would lead to occlusion problems, and the general trends of gradient can be viewed in the average.

It may also not be necessary to step all the way back through the batch when calculating itemized gradients.  Equation (\ref{eqn:m_decomp}) shows that the gradient due to a particular time step's loss is decomposable into a sum of sequence.  We can use (\ref{eqn:m_ratio}) to analyze the rate of decay of that sequence.

\begin{align}
\frac{M_j}{M_{j+1}} = \frac{h_{j-1}}{h_j}W\tanh'_{j}
\end{align}

$W$ is generally initialized close to 0, and regularization is used to keep it having small magnitude during gradient descent.  $\tanh'$ has a range of $(0,1]$, and $\frac{h_{j-1}}{h_j}$ should generally be close to one.  Thus, the sequence $\{M_j\}$ should decay as $j$ decrements.  Then it would stand to reason that we might be able to choose a value $k$ such that we only have to step back $k$ steps to be close enough to the real gradient.

\begin{align}
S_k &= M_i + M_{i-1} + \ldots + M_{i-k}\\
S_k &\approx \frac{\partial h_i}{\partial W}
\end{align}

In our use case, we empirically chose $k = 5$ based on manual inspection of the data.  However, it is highly likely that acceptable $k$ should vary with RNN architecture and cell choice.  It might be possible to find a $k$ such that $||S_k - \frac{\partial h_i}{\partial W}|| < \epsilon$ globally across a validation set.  Since $k$ is loosely a measure of how far the gradient horizon is, it would stand to reason that a more sophisticated architecture would demand a larger $k$.

\section{Use Case}
\label{sec:use_case}

To demonstrate the use of RNNbow, we trained a character-level RNN on the Linux Kernel to try to get it to generate code that looks like the $C$ programming language, replicating an experiment done in Karpathy et al.'s seminal RNN work\cite{DBLP:journals/corr/KarpathyJL15}.  We used batches of 25 characters, and recorded gradients every 100 batches over the first 50000 batches of the training data.  We use a hidden layer of 100 nodes, but we average the gradients across all nodes.  In this section, we outline several insights that can be found via RNNbow.

\begin{figure}[ht!]
  \centering
  \centering
  \includegraphics[width=0.4\textwidth]{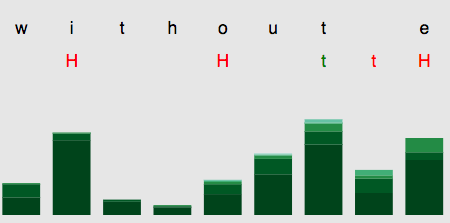}
  \caption{The gradients of a batch early in training.  Note that the gradients are
  mostly composed of darker shades of color.  This signifies that the gradient updates are primarily due to local loss, zero, one or two time steps away.}
  \label{ref:small_horizon}
\end{figure}
\begin{figure}[ht!]
  \centering
  \includegraphics[width=0.4\textwidth]{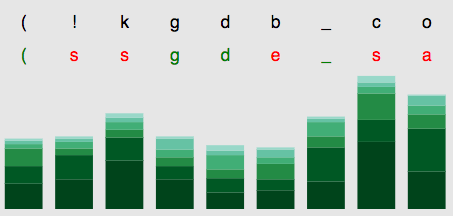}
  \caption{The gradients of a batch later in training.  Here, the gradients are much more distributed across different shades, suggesting that longer-term dependencies are being learned.}
  \label{ref:long_horizon}
\end{figure}

\subsection{Overview of Gradients Over Time}

Figure~\ref{ref:interface} shows the result of training an RNN using our approach. Looking at the overview section, seen as area 1 in Figure~\ref{ref:interface}, the first insight is that the magnitude of the gradient starts very small, and then appears to plateau, albeit with a fair amount of variance.  This suggests that early in training, the weights update slowly - there may be some burn-in required before the parameters are updating efficiently.  The overview also points the user to batches with maximal gradient.  It makes it easy for the user to view the elements of the training set that the RNN learns the most from.  

It can also be instructive to compare the batch visualizations (area 2 in Figure~\ref{ref:interface}) as they change from early in training to late in training.  Figure~\ref{ref:small_horizon} shows the gradients of a batch early in training, and Figure~\ref{ref:long_horizon} shows the gradients of a batch late in training.  At a glance, the darker the batch visualization is, the shorter the gradient horizon is; a larger portion of the update at each step comes from local losses.  In Figure~\ref{ref:small_horizon}, most of the bars in the visualized batch are primarily composed of dark green bars.  Compare that pattern to a batch later in training in Figure~\ref{ref:long_horizon}, where the gradient is much lighter; this corresponds to longer gradient horizons for training in this batch.  Exploring the training patterns over training reveals that this particular RNN seemed to lengthen its time dependencies as training went on.

\begin{figure}[ht!]
 \includegraphics[width=0.5\textwidth]{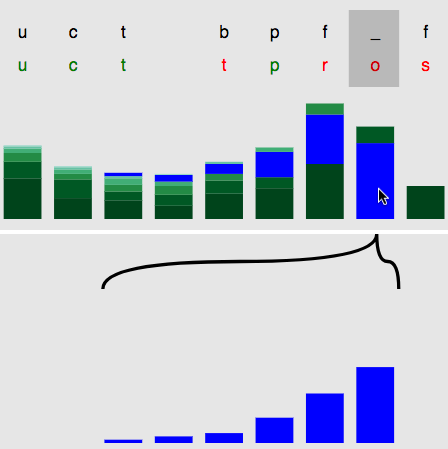}
 \caption{The detailed view of the gradients due to a single character's loss.  Note that it makes the vanishing gradient effect very apparent.}
 \label{img:vanishing_gradient}
\end{figure}

\subsection{Vanishing Gradient}
\label{sec:vanishing_gradient}

% Consider the sentence \textit{The }\textbf{man} \textit{receieved a phone call on the way to} \textbf{his} \textit{car.}  In order for an RNN to be able to know the gender of the pronoun \textbf{his}, it must remember the gendered noun 9 time steps earlier.  Since RNNs learn via gradient descent, the only way to learn time dependencies of that distance is to have the gradient due to a loss propagate back a far distance.  If the word \textbf{his}, at $t=i+9$, is a function of the word \textbf{man}, at $t=i$, then $\frac{\partial y_{i+9}}{\partial h_{i}}$ must be greater than 0.

A well-known consequence of the activation functions used in RNNs, $\tanh$ and $\sigma$, is that they result in a gradient that decreases as it is passed back in time.  For large swaths of the hyperparameter space, the gradient may decay incredibly fast, restricting any long-term dependency learning\cite{Pascanu:2013:DTR:3042817.3043083,Hochreiter:1998:VGP:353515.355233}.  

The primary function of the projection of gradients in the visualization, area 4 in Figure \ref{ref:interface}, is to illustrate the rate at which the gradient decays.  By mousing over a gradient bar, the user can see the rate at which that particular gradient due to a particular character's loss vanishes.  This can be seen in Figure \ref{img:vanishing_gradient}.

\begin{figure}[ht!]
 \includegraphics[width=0.5\textwidth]{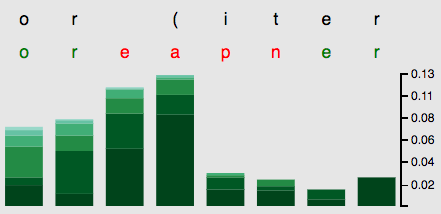}
 \caption{The maximal gradient in the training set.  The RNN assumes a large gradient when predicting the character \texttt{a} instead of the character \texttt{(}.  This may be due to a large loss being incurred by the model not learning the iterator grammar of the \textit{C} programming language.}
 \label{img:maximal_gradient}
\end{figure}

\subsection{Batches With Maximal Gradient}

The overview bar chart in area 1 of the interface shown in Figure~\ref{ref:interface} can be used to cue the user towards parts of the training set that the RNN learns the most from.  For an example, we clicked on the rightmost bar of area 1 to change the focus of RNNbow to that batch, since that bar had the greatest height and thus the greatest maximal gradient.  The stacked bar chart of the maximal gradient in that batch can be seen in Figure~\ref{img:maximal_gradient}.  The maximal gradient is due to predicting the character \texttt{a} instead of predicting the character \texttt{(}, in spite of the context of being in a \texttt{for} loop.  Note that it also assumes some gradient from incorrectly predicting the subsequent characters as well.  This suggests that our RNN has not learned the iterator grammar of the \textit{C} programming language.  It also confirms that our RNN is learning from actual mistakes and not overfitting the training set.  As this maximal gradient comes late in our training set and is on a legitimate mistake, it cues the user that we have not trained enough and training must continue.

\section{Discussion}

\subsection{Limitations}

RNNbow is not intended to be a tool for power users of RNNs; such users would be better served with custom visualizations and custom analytics within their deep learning pipelines.  RNNbow is most useful to the non-expert.  The current implementation does have some scaling issues, both in the interface as well as in the implementation of backpropagation through time, described in section~\ref{sec:bptt}.  However, it is more likely that a non-expert would use smaller networks that are executable on a personal computer; that is the scale we aim to currently support.

The design does have some visual layout issues with scale as well.  This interface doesn't support more than a few hundred batches, although this should be solvable with some aggregation and drilldown.  In practice, RNNs may train over hundreds of thousands of batches.  A heuristic could be used to point the user to particularly interesting batches within the training data.  Similarly, the stacked bar chart might not scale to batch sizes of 50 or more.  In the use case given in this work, with a batch size of 25 and $k=5$, the stacked bar chart was responsible for visualizing 125 pieces of data.  A sophisticated RNN might have a batch size of 128 and would hopefully have a much larger memory; more iterations of design are needed to come up with an appropriate visualization of so many gradients.  In addition, some form of aggregation would need to be defined for the predicted labels (in this case, characters) in large batch sizes.  Perhaps the largest hurdle to supporting industry scale networks is the number of layers visualized.  RNNbow currently supports a single layer; there are popular CNNs with more than a hundred layers, and RNNs are following suit\cite{DBLP:journals/corr/PascanuGCB13}.  It's unclear how the stacked bar chart would scale to even a dozen layers.  It may be that a higher resolution visualization of gradients between layers is needed.

Prior to the design of the visualization, basic experiments were run to estimate the gradient horizon throughout the first 50000 batches of training, and it was found that it varied from 2 to 5 characters before the gradient dropped below a certain $\epsilon > 0$.  Production-level RNNs, with sophisticated architectures, may need gradient horizons of tens or even hundreds of time steps.  The current algorithm used may not be scalable to record the gradient that many steps back.  This could be addressed by taking the gradients much less frequently than every 100 batches.

\subsection{Visualizing Gradients}

One of the key insights of RNNbow is that it visualizes the gradient, as opposed to the activation.  Gradients are increasingly being accepted as a more salient representation of the learning surfaces and thus a more informative value to study than activation and loss\cite{koh2017understanding}.  Considering that the gradient is what dictates updates to the model, i.e. what is learned, its visualization is revealing of the training process.  It is particularly salient in a visualization of RNNs, as the many-to-many relationship between losses and time steps can be difficult to intuit about.  It proved useful in discovering endemic properties like vanishing gradients; there may be other endemic qualities in network training that cannot be noticed in visualizations of activations.  

In this work, we focused on the gradient $\frac{\partial L_i}{\partial W}$, in part because this is a value that is already traditionally calculated during backpropagation, as it represents the quantity by which we update the tunable parameter $W$ in the RNN.  This in turn led to calculating intermediate gradients including $\frac{\partial L_i}{\partial y_i}$ and $\frac{\partial h_i}{\partial W}$.  There are other gradients, however, that could have been calculated during training.  In particular, $\frac{\partial h_i}{\partial h_j}$ could have revealed interesting interplay between nodes within the hidden layer and how they kept track of memory.  As we seek to crack open the black box of neural networks, it is important to remember that each gradient is a different tangent surface to the manifold on which the full parameter space of the network lies.  In this work, we analyze a very specific tangent surface with the goal of understanding time dependencies.  In work on visualizing CNNs, it is very common to consider gradients of the activations as they correspond to particular outcomes with a goal of determining what parts of the image are most sensitive to a network's understanding of a label.  In order to better understand the cross effects between parameters in a deep network, it may be necessary to explore several of these gradient surfaces, coupled with a visual analytics system that guides the user towards logical conclusions, as RNNbow does for the vanishing gradient.  Each gradient surface may change drastically over different architectures and settings of hyperparameters.  Comparing gradient surfaces generated by different networks may offer more insight into the sensitivity of network behavior to each hyperparameter.

\subsection{Role of Visual Analytics in Deep Learning}

It is not unreasonable to ask: \textit{is there even a role for Visual Analytics in Deep Learning}?  While it is clear that visualization has a role in helping a user build models that they trust, there are extenuating circumstances that make it hard to apply the same considerations to deep learning.  A typical Support Vector Machine or Decision Tree has only a handful of hyperparemeters that need to be set by the user, and it learns only a dozen or so parameters during training.  In contrast, the famous AlexNet CNN from 2012\cite{NIPS2012_4824}, a relative dinosaur in deep learning years, fits 60 million parameters.  Users must choose architectures, cell types, learning rate, filter size, and a number of other hyperparameters.  While this may sound like it offers a good opportunity for building an analytics system that abstracts away these decisions, in practice, it is impractical to infer these values.  Generally, visual analytics systems that collaboratively build a model with a user do so by engendering an abstraction, or mental model, of the underlying parameters, and then enable the user to manipulate those parameters with intuitive interactions.  In many cases, industry-size neural networks simply have too many underlying parameters, and these parameters tend to interact with one another in unintuitive ways.  Once a user has trained enough to understand how those parameters work together, they likely have gained the ability to program their own scripts using one of the many fully-featured deep learning frameworks, at which point they may find a visual analytics sytem superfluous.

Perhaps because most builders of deep learning models were already able to write their own scripts, the most succesful visualizations related to deep learning were not incorporated into visual analytics systems, but were rather used to try to explain the inner workings of a deep learning model post-hoc, after it had been built, in order to gain insight into some of the structures that empirically were working.  This led to static visualizations of cleverly calculated values, including gradients, typically overlaid on the input domain to ground the network in the data on which it had been trained.  In their popularity, these visualizations suggested that it could be possible to apply some interactivity into model building if the \textit{right} data was visualized.

The current usage of deep learning in the wild This landscape leads to very different recommendations for designing tools depending on the audience of the tool.  Our tool is primarily targeted at nonexperts tuning simple one-layer RNNs, possibly in an educational setting.  For that reason, we favored \textit{simple, low-dimensional visualizations}, and used \textit{calculated values with a clear, unambiguous meaning}.  For an expert, the design decisions are drastically different.  The designer can make an assumption that the user will be directly interacting with their model through a deep learning scripting framework and that this framework gives them the ability to calculate their own values of interest.  Then these works can focus more on an overview of many different aspects of the network, almost like a monitoring system.  Good examples of support for industry-scale model builders can be found in recent works by Google\cite{wongsuphasawat2018visualizing} and Facebook\cite{kahng2018cti}.  For a more thorough treatment of the state of visual analytics in deep learning, we direct the reader to a recent survey\cite{hohman2018visual}.

\section{Future Work}

The design space of RNNs is rich and constantly evolving, and much of the progress has the goal of having a longer gradient horizon.  As visualizing this gradient horizon is the key feature of RNNbow, it should prove invaluable in comparing different approaches.  One direction of RNN research focuses on making more sophisticated cells in which the calculation of the hidden state and outputs differs from the vanilla RNN equations, seen in (\ref{eqn:hidden_state}) and (\ref{eqn:output_calc}).  We would like to adapt RNNbow to visualize the gradient flow of these alternative cells.  It should be illustrative in comparing a vanilla RNN to a Long Short Term Memory (LSTM).  LSTMs have two hidden layers, one of which is supposed to hold short term memory, and one of which holds long term memory.  It would be interesting to see if this is supported in visualizations of the gradients of each hidden layer. Another type of cell called a Gated Recurrent Unit (GRU) only has one hidden layer, but in practice accomplishes long-term dependencies similarly to LSTMs.  It isn't particularly clear why this should be.  In both LSTMs and GRUs, the calculations of emissions require additional computation with each type of cell having several gates that supposedly lengthen the memory.  Perhaps RNNbow would be able to be extended to reveal the gradient flow within a cell, not just between cells.

Beyond different cell types, different arrangements of cells have also proved helpful in practical RNNs.  One type of cell architecture is a bidirectional RNN, in which two recurrent neural networks are run in parallel for each batch, with one running through the input sequence from left to right and the other in reverse.  Their outputs are then combined through a learned linear combination.  The bidirectional RNN allows learning former and future context, and viewing the gradient flow in both directions may aid in the understanding of its training.  More advanced architectures results from adding layers of RNNs, either to match to multidimensional sequences, or to use multiple layers to capture different levels of abstraction in the input sequence as is typical in CNNs.  These architectures are difficult to conceptualize; visualizing the gradient may help.  However, their spatial complexity proves a challenge in the current layout of RNNbow; some cleverness will be necessary to determine a layout for such architectures.  

In this work, we didn't consider the separate nodes in the hidden layer - we just averaged the gradients together.  This is in contrast to many of the works on interpreting ANNs, in particular because the activation of a single node tends to be a binary decision maker.  Previous work suggests that individual nodes have unique responsibilities; in the same experiment as used in this work, Karpathy et al.\cite{DBLP:journals/corr/KarpathyJL15} found that a single node was responsible for remembering the state of generated $C$ code as being in a parenthesis or out of a parenthesis.  However, there is no scalable way to visualize all hidden nodes in RNNbow; some heuristic will need to be developed to cue the user to interesting nodes, and some overview of node performance other than the mean will need to be added.

\section{Conclusion}

We present RNNbow, a web-based visualization tool for analyzing gradient flow during training of a Recurrent Neural Network.  We demonstrate how it can be used to find endemic properties in a network, and how it can provide insights into the learning process.  We also show that it can be a useful educational tool for illuminating the vanishing gradient phenomenon.  We review how to calculate the itemized gradients necessary for loading data into RNNbow.  We discuss potential uses of the tool, especially the applications to other RNN architectures.

%% if specified like this the section will be committed in review mode
\acknowledgments{
This material is partially based on research sponsored by the Air Force Research Laboratory and DARPA under agreement number FA8750-17-2-0107.  This work is also supported in part by the National Science Foundation: IIS-1452977 (CAREER). }

% \Urlmuskip=0mu plus 1mu
%\bibliographystyle{abbrv}
\bibliographystyle{abbrv-doi}
%\bibliographystyle{abbrv-doi-narrow}
%\bibliographystyle{abbrv-doi-hyperref}
%\bibliographystyle{abbrv-doi-hyperref-narrow}
% \raggedright
% \sloppy
\bibliography{rnnvis}

\appendix

\section{Itemized Gradients vs. Backpropagation}

The calculation of (\ref{eqn:full_grad_double_sum}) takes $O(n^2)$, where $n$ is the size of the batch.  It is possible to utilize dynamic programming further to speed it up to $O(n)$; this is used in most implementations of backpropagation.  First, we expand both sums in (\ref{eqn:full_grad_double_sum}).

\begin{align}
\frac{\partial L}{\partial W} = &\frac{\partial L_n}{\partial y_n} \cdot \frac{\partial y_n}{\partial h_n} \cdot\left[M_n + M_{n-1} + \cdots + M_1\right] \nonumber \\
& + \frac{\partial L_{n-1}}{\partial y_{n-1}} \cdot \frac{\partial y_{n-1}}{\partial h_{n-1}} \cdot \left[M_{n-1} + \cdots + M_1\right] \nonumber \\
& \cdots \nonumber \\
& + \frac{\partial L_{1}}{\partial y_{1}} \cdot \frac{\partial y_{1}}{\partial h_{1}} \cdot  M_1
\end{align}

Next, we distribute, group the terms by $M_j$, and factor.

\begin{align}
\label{eqn:distributed_backprop}
\frac{\partial L}{\partial W} = &M_n \left(\frac{\partial L_n}{\partial y_n} \cdot \frac{\partial y_n}{\partial h_n}\right)\nonumber \\
+ &M_{n-1} \left(\frac{\partial L_n}{\partial y_n} \cdot \frac{\partial y_n}{\partial h_n} + \frac{\partial L_{n - 1}}{\partial y_{n - 1}} \cdot \frac{\partial y_{n - 1}}{\partial h_{n - 1}}\right)\nonumber \\
+ &\cdots \nonumber \\
+ &M_1 \left(\frac{\partial L_n}{\partial y_n} \cdot \frac{\partial y_n}{\partial h_n} + \ldots + \frac{\partial L_1}{\partial y_1} \cdot \frac{\partial y_1}{\partial h_1}\right)
\end{align}

Let $N_j = \left(\frac{\partial L_n}{\partial y_n} \cdot \frac{\partial y_n}{\partial h_n} + \ldots + \frac{\partial L_j}{\partial y_j} \cdot \frac{\partial y_j}{\partial h_j}\right)$.  Then $\{N_j\}$ can be implemented with a dynamic programming table as with $\{M_j\}$, and we can calculate the gradient in a single pass.

\begin{align}
\label{eqn:backprop}
\frac{\partial L}{\partial W} &= \sum_{i=1}^n M_i N_i\\
N_j &= N_{j+1} + \frac{\partial L_j}{\partial y_j} \cdot \frac{\partial y_j}{\partial h_j} \hspace{5mm} ; \hspace{5mm} 0 < j < i \\
N_i &=  \frac{\partial L_i}{\partial y_i} \cdot \frac{\partial y_i}{\partial h_i}
\end{align}

In optimized versions of backpropagation through an RNN, we only make a single pass backwards through time, passing back our accumulation of the gradients $N_j$, and adding on the gradient of the current time step.  For RNNbow, we can't use this method, however, because we lose track of the terms in the expanded product of (\ref{eqn:distributed_backprop}) when we make use of the dynamic programming table for $\{N_j\}$.  Thus, we need to use the $O(n^2)$ version described by (\ref{eqn:full_grad_double_sum}), saving each summand as we accumulate the sum.  It is possible that, depending on the implementation library, keeping track of the intermediate $M_j$ and $N_j$, and then utilizing vector math, as is commonly used in the python library Numpy, could allow us to use traditional backpropagation.

\end{document}